\documentclass[runningheads]{llncs}

\usepackage{eccv}

\usepackage{eccvabbrv}

\usepackage{graphicx}
\usepackage{booktabs}
\usepackage{algorithm}
\usepackage{algorithmic}
\usepackage{subcaption} % Required for subfigures
\usepackage{amsmath}
\usepackage{amssymb}
\usepackage[accsupp]{axessibility}  % Improves PDF readability for those with disabilities.

\usepackage{hyperref}

\usepackage{orcidlink}

\begin{document}

\title{FedLAS: Feature-Modulated Bidirectional Label Smoothing for Neural Network Calibration} 
%\title{\name : Feature-Dependent Label Smoothing for Dynamically Rectifying Over- and Under-confidence in Deep Neural Networks} 

\titlerunning{FeDLaS}

% \author{Thiru Thillai Nadarasar Bahavan \inst{1}\orcidlink{0000-1111-2222-3333} \and
% Sachith Seneviratne\inst{2,3}\orcidlink{1111-2222-3333-4444} \and
% Saman Halgamuge\inst{3}\orcidlink{2222--3333-4444-5555}}

% % TODO FINAL: Replace with an abbreviated list of authors.

% % TODO FINAL: Replace with your institution list.
% \institute{University of Melbourne, Parksville VIC 3053, Australia} \\
% %\url{http://www.springer.com/gp/computer-science/lncs} \and
% % ABC Institute, Rupert-Karls-University Heidelberg, Heidelberg, Germany\\
% % \email{\{abc,lncs\}@uni-heidelberg.de}}

\author{
Thiru Thillai Nadarasar Bahavan\inst{1}\orcidlink{0000-0001-7437-3609} \and
Sachith Seneviratne\inst{1}\orcidlink{0000-0001-9094-2736} \and
Saman Halgamuge\inst{1}\orcidlink{0000-0002-2536-4930}
}

\authorrunning{Bahavan et al.}

\institute{
The University of Melbourne, Parkville, Australia\\
\email{bahavant@student.unimelb.edu.au, \{sachith.seneviratne,saman.halgamuge\}@unimelb.edu.au}
}

\maketitle
\newcommand{\nameplug}{FeDLaS}
\newcommand{\namels}{FeDLaS-LS}
\newcommand{\namembls}{FeDLaS-MbLS}

\begin{abstract}
Deep Neural Network (DNN) classifiers suffer from poor calibration when their softmax outputs (predictive confidence) deviate from the empirical likelihoods. This manifests itself as either overconfident incorrect predictions or under-confident correct predictions. Label smoothing (LS) enhances model calibration by introducing entropy regularization during training through redistributing probability mass from the ground-truth label to the remaining classes. LS, including Margin-based LS (MbLS), have restrictive assumptions: they rely on predefined, uniform smoothing rules and only tackle overconfidence. In reality, samples exhibit diverse characteristics, such as difficulty/ambiguity, that interact with the evolving nature of the model being trained. In training, samples may have various degrees of under- or overconfidence. To overcome this, a mechanism that identifies the specific confidence state of each sample and determines the appropriate degree of smoothing in each training step is needed, tailoring the adjustment to the individual sample. We propose \nameplug: Feature-Modulated Bidirectional Label Smoothing, a plug-and-play algorithm for label smoothing-based losses. In \nameplug, we introduce a Feature Norm-based Confidence Indicator (NCI) to control smoothing and a Bidirectional Calibration Gating (BCG) module to detect both over and under-confidence. Our algorithm can be integrated with LS and MbLS based losses when applied to standard DNNs, enhancing performance. Extensive experiments on standard and fine-grained high-resolution vision benchmarks show that \nameplug~consistently improves calibration compared to modern baselines, reducing Expected Calibration Error (ECE) and Adaptive ECE while maintaining Top-1 accuracy. Code: github.com/nadarasarbahavan/FEDLAS
\end{abstract}

\section{Introduction}
\label{sec:intro}

Reliable confidence estimation is fundamental to decision-making; \eg, in clinical diagnostics, it determines whether the prediction of a DNN classifier can be trusted for treatment or whether the case must be escalated to a human specialist. A DNN is considered well-calibrated when its predicted confidence estimates align with the true empirical likelihoods. However, a significant issue arises with many over-parameterized DNN-based classifiers: they are often poorly calibrated~\cite{guo2017calibration}, meaning their predicted probabilities fail to accurately reflect the true likelihood of correctness. For example, if the clinical diagnostics classifier assigns 70\% confidence to a set of predictions, ideally 70\% of those predictions should be accurate. 

\begin{figure*}[t]
    \centering
    % --- Subfigure 1: Sigmoid Saturation ---
    \begin{subfigure}[b]{0.2\textwidth}
        \centering
        \includegraphics[width=\linewidth]{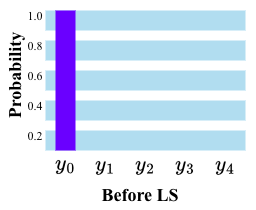}
        \caption{One-hot Label}
        \label{fig:lsbefore}
    \end{subfigure}
    \hfill
    \begin{subfigure}[b]{0.2\textwidth}
        \centering
        \includegraphics[width=\linewidth]{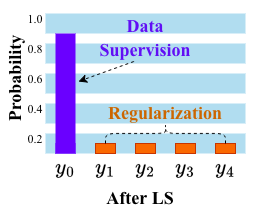}
        \caption{LS Label}
        \label{fig:lsafter}
    \end{subfigure}
    \hfill
    % --- Subfigure 2: EMA vs Batch ---
    \begin{subfigure}[b]{0.2\textwidth}
        \centering
        \includegraphics[width=\linewidth]{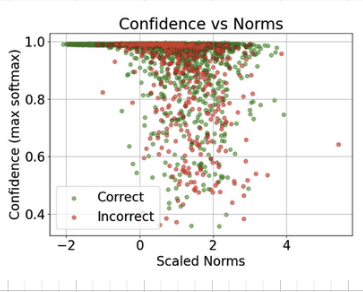}
        \caption{$L_1$ vs $\text{max}(p_i)$}
        \label{fig:ema_stability}
    \end{subfigure}
    \hfill
    % --- Subfigure 3: Norm Distribution ---
    \begin{subfigure}[b]{0.32\textwidth}
        \centering
        \includegraphics[width=\linewidth]{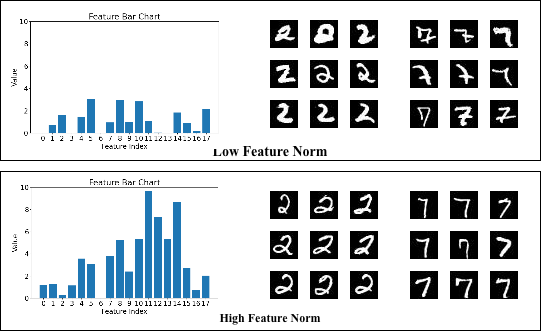}
        \caption{Low vs High $L_1$}
        \label{fig:norm_dist}
    \end{subfigure}
    
    \caption{(a)-(b) Labels before and after Label Smoothing. (c) For each sample, softmax probability values (X-axis) tends to saturate towards very high values, with little variation. In contrast, $L_{1}$ captures more of the variation. (d) The features reflect the activation tendencies of the easy-to-learn and hard-to-learn images. Spurious difficult images have a flatter features, resulting in a low $L_{1}$ values. The activation tendency can thus be summarized via its $L_{1}$ norm.}
    \label{fig:mnist_norms}
\end{figure*}

In standard DNN training, networks maximize the likelihood of training labels. The use of one-hot encoded labels (\cref{fig:mnist_norms}(a)) encourages the models to assign near-certain probabilities to the correct class~\cite{guo2017calibration}, forcing the predictions to closely match the ground truth. This leads to overfitting and causes models to exhibit overconfidence, even when predictions are incorrect~\cite{guo2017calibration}. Label smoothing (LS) improves model calibration by introducing entropy regularization during training by redistributing the probability mass from the ground-truth label to the remaining classes (\cref{fig:mnist_norms}(b)). LS, including Margin-based LS (MbLS) have restrictive assumptions: they rely on predefined, uniform smoothing rules and only tackle overconfidence~\cite{mbls,ls}. In reality, samples exhibit diverse characteristics, such as difficulty/ambiguity, that interact with the evolving nature of the model being trained. In training, samples may have various degrees of under- or overconfidence. To overcome this, \textit{a mechanism that identifies the specific confidence state of each sample and determines the appropriate degree of smoothing in each training step is needed, tailoring the adjustment to the individual sample}.

Park et al.~\cite{acls} mathematically demonstrated that many advanced network calibration methods are implicitly sample-adaptive variants of label smoothing that rely on softmax output to determine the degree of smoothing~\cite{mmca,ecp,MDCA,CPC,CRLcheck,focalloss,mbls}. The authors claim that these methods have inherent structural limitations acknowledged by \cite{acls}, \eg, \textit{Lim1}: these methods do not correct calibration across the complete range of predictive confidence. However, three more critical challenges (\textit{Lim2-Lim4}) remain unaddressed: \textit{Lim2:} Most of these methods remain constrained by a unidirectional focus on overconfidence, overlooking scenarios where samples exhibit under-confidence. Some methods account for under-confidence, but they use a validation dataset~\cite{adafocal}. \textit{Lim3:} They do not accommodate the non-stationarity of the model when training. \textit{Lim4:} They use softmax outputs that are naturally prone to be overconfident. Motivated by these four limitations, we address the following two research questions: 1. During training, how can the direction of miscalibration (under-confidence vs. over-confidence) be identified for each sample, without the use of auxiliary data? 2. Beyond the model output, can we leverage the models' internal behavior to address model miscalibration? Our contributions are as follows.

\begin{itemize}
    \item \textbf{Norm-based Confidence Indicator:} To address \textit{Lim3} and \textit{Lim4}, we introduce NCI, a label-free indicator leveraging hidden-layer feature norms. Feature norms are an established class-agnostic proxy of predictive confidence~\cite{featurenorm, vaze, sphor, adaface}, and theoretically the confidence of the network's hidden classifier~\cite{featurenorm}; we ask \textit{whether they can also correct network miscalibration}. NCI bypasses softmax overconfidence and normalizes these norms against an Exponentially Moving Average (EMA) reference to handle non-stationarity.
    \item \textbf{Bidirectional Calibration Gating (BCG):} To address \textit{Lim2}, we propose the BCG module, a dynamic mechanism designed to detect and mitigate both overconfidence and under-confidence pointwise throughout the training process, without a validation dataset. 
    \item \textbf{Adaptive Smoothing Module (ASM):} By combining NCI and BCG, we develop the ASM, which adjusts label smoothing per-sample across the entire spectrum of predictive confidence, addressing \textit{Lim1}. We demonstrate how to integrate it into Vanilla and Margin-based Label Smoothing.
    \item \textbf{Extensive Empirical Validation:} We conduct comprehensive experiments across a diverse set of benchmarks, including CIFAR-10/100, Tiny-ImageNet, two high-resolution fine-grained classification tasks. Our results consistently demonstrate superior performance in both accuracy and model calibration.
\end{itemize}

\section{Related Work}
\label{sec:related}
\textbf{Post hoc approaches} Post-processing methods offer a simple and efficient way to calibrate predictions by transforming the output of a model \cite{guo2017calibration, zhang2020mix, Tomani2021Posthoc}. Temperature scaling \cite{guo2017calibration}, a streamlined variant of Platt scaling \cite{largemargin}, uses a single parameter to soften pre-softmax logits. Although effective in-domain, it performs poorly under distributional shift \cite{ovadia2019can}.
\textbf{Probabilistic and non-probabilistic methods.} Probabilistic methods include Bayesian neural networks \cite{blundell2015weight, louizos2016structured}, stochastic expectation propagation \cite{hernandez2015probabilistic}, and dropout-based variational inference \cite{gal2016dropout}. Ensemble learning, a popular non-parametric alternative, estimates uncertainty via the empirical variance of predictions and improves calibration by aggregating outputs from diverse models. Common strategies involve varying hyperparameters \cite{wenzel2020hyperparameter}, random seeds and data shuffling \cite{lakshminarayanan2016simple}, Monte Carlo Dropout \cite{gal2016dropout}, modeling dataset shift \cite{ovadia2019can}, and enforcing model orthogonality \cite{larrazabal2021orthogonal}. Despite their effectiveness, ensembles are computationally expensive, particularly with large models and datasets.

\textbf{Explicit and implicit penalties.} These methods introduce entropy-based penalties during training to regularize predictions and mitigate overconfidence. ECP \cite{ecp} and label smoothing \cite{ls} explicitly maximize entropy by penalizing overconfident outputs. Focal loss \cite{focalloss} implicitly increases entropy by minimizing the KL divergence between softmax outputs and a uniform distribution \cite{calibfocal}. As shown in \cite{ls,calibfocal,dfl}, both act as implicit regularizers that promote uncertainty. 
On a different track, focal loss based methods have been put forth to handle both over and under-confidence: such as Dual Focal Loss and Adaptive Focal Loss. AdaFocal uses a validation set, which is limiting and DFL applies regularization to limited range of predictive confidence. MbLS relaxes the entropy constraint \cite{mbls}, and ACLS\cite{acls} builds on MbLS by adapting it to overconfidence \cite{acls}. Overall, loss functions that encourage higher prediction entropy, explicitly or implicitly, has been shown to achieve state-of-the-art calibration performance \cite{ls,calibfocal,acls}.

\textbf{Other Methods} Tao introduced feature clipping as a method for applying an entropy-based penalty \cite{Tao2025FeatureClipping}. Earlier works, such as \cite{prob}, have investigated leveraging features to group similar samples in pursuit of multi-calibration objectives \cite{HbertJohnson2018MulticalibrationCF}. However none of these explored it for adaptive label smoothing. There are Mixup based methods such as RankMixup, however this involves significant data augmentation~\cite{rankmixup}.  

\textbf{Feature Norms} Recently, various variants of feature norms has been shown (both empirically and theoretically) to differentiate via ordering between in-distribution and out-of-distribution samples \cite{featurenorm, vaze, sphor},  low quality and high quality images \cite{adaface} and data with high and low uncertainty for metric learning \cite{vonmisesloss}.

\section{Preliminaries \& Problem Formulation} \label{premsprobform}
\subsubsection{Setup.}
The training dataset $S = \{(\mathbf{x}^{(i)}, \mathbf{y}^{(i)})\}_{i=1}^N$ consists of $N$ input-label pairs. Each input $\mathbf{x}^{(i)} \in \mathcal{X}$ is associated with a ground truth label $y^{(i)}$ from a label space $\mathcal{Y}$ consisting of $\{1,...,C\}$ classes. The label $y^{(i)}$ is encoded as a one-hot vector $\mathbf{y}^{(i)}$. Specifically, for a sample $i$ with true class label $j$, the label is defined as $\mathbf{y}^{(i)} = \{y_{k}^{(i)}\}_{k=1}^{C}$, where $y_{j}^{(i)}=1$ and $y_{k}^{(i)}=0$ for all $k \neq j$. We define a neural network classifier $f_\omega(\mathbf{x})$ trained on $S$ that maps an input to a probability distribution $\hat{\mathbf{p}}(\mathbf{x}) \in  \Delta^{C-1}$. The model is typically trained by minimizing the \textit{cross-entropy (CE) loss} $\mathcal{L}_{\text{CE}} = \frac{1}{N}\sum_{i=1}^{N}\mathcal{L}^{i}_{\text{CE}}$, where $\mathcal{L}^{i}_{\text{CE}} = -\sum_{k=1}^{C} y^{(i)}_k \log \hat{p}^{(i)}_k$.

\subsubsection{Problem statement.}
From the predicted distribution, the classifier provides a prediction index $i = \arg\max_{k \in \{1,\cdots,C\}} \hat{p_k}^{(i)}$, with an associated \textbf{predictive confidence} $\hat{p}_c =\max_{k \in \{1,\cdots,C\}} \hat{p_k}^{(i)}$. A model is \textbf{perfectly calibrated} if this confidence $\hat{p}_c$ represents the true likelihood of correctness $p$. Formally, a model is perfectly calibrated if:
\begin{equation}
    P(\hat{{y}} = {y} \mid \hat{p}_c = p) = p, \quad \forall p \in [0, 1]
\end{equation}

The miscalibration manifests itself as a gap between the predicted $\hat{p}_c$ and $p$, specifically as overconfidence $P(\hat{y} = y \mid \hat{p}_c = p) < p$ or under-confidence $P(\hat{y} = y \mid \hat{p}_c = p) > p$. The primary goal of \textit{calibration methods} is to minimize this disparity. This disparity is measured via Expected Calibration Error (ECE) and Adaptive Expected Calibration Error (AECE)~\cite{mbls,ece,eace}.

\subsubsection{Motivation for Feature norm as Confidence proxy.}
Following previous work, we quantify the predictive confidence of the model by examining its internal behavior, using the $L_1$ norm of the $d$-dimensional feature vector $\mathbf{z}^{(i)} = F_{backbone}(\mathbf{x}^{(i)})$. Generally, $L_p$ norms are defined as $\|\mathbf{z}^{(i)}\|_p = \left( \sum_{j=1}^{d} |z_j^{(i)}|^p \right)^{1/p}$ for $p \geq 1$.

% For a $d$-dimensional vector $\mathbf{z}$, its $L_p$ norm is $\|\mathbf{z}\|_p = \left( \sum_{j=1}^{d} |z_j|^p \right)^{1/p}$, $p \geq 1$. We use the $L_1$ norm of the feature vector $\mathbf{z}^{(i)} = F_{\text{backbone}}(\mathbf{x}^{(i)})$, since feature-vector norms have established functional utility as a measure of predictive confidence~\cite{featurenorm, vaze, sphor, adaface}. This motivates one of our method's components; we investigate whether it extends to network calibration.

To provide a theoretical basis for this approach, we draw upon the findings of Park et al.~\cite{featurenorm}. They demonstrate that for networks utilizing unit-wise rectifier activation functions (\eg, ReLU or GeLU), the $L_1$ norm of a hidden layer's feature vector acts as the maximum logit of an implicit \textit{hidden classifier}\cite{featurenorm}. The logit $s(x)$ is the pre-softmax output of the classifier. For notational consistency with \cite{featurenorm}, let activations in layer $(l)$ be denoted as $\mathbf{a}^{(l)}$, which corresponds to our extracted feature vector $\mathbf{z}^{i}$, and let logits be $\psi(x):=s(x)$.

\noindent \textbf{Theorem (Park et al., 2023)\cite{featurenorm}.} \textit{The final logits of the model are represented by the linear transformation: $\mathbf{\psi}(\mathbf{x}) = \mathbf{C}^{(l)}\mathbf{a}^{(l)}$, where $\mathbf{C}^{l}$ is the derived coefficient matrix and $\mathbf{a}^{(l)}$ represents the activations at layer $l$. A binarized hidden classifier $\overline{\mathbf{\psi}}^{l} \in \mathbb{R}^C$ can be derived such that $\overline{\mathbf{\psi}}^{l}(\mathbf{x}) := \text{sign}(\mathbf{C}^{l})\mathbf{a}^{(l)} = \mathbf{B}^{l}\mathbf{a}^{(l)}$. Under sufficient discriminative training for class $j$, the feature norm $\|\mathbf{a}^{l}\|_1$ converges to the hidden confidence $\overline{\psi}^{l}_j(\mathbf{x})$ (Logit of class $j$). Specifically, for any class $j$:}
\begin{equation}
0 \leq \|\mathbf{a}^{(l)}\|_1 - \overline{\psi}^{l}_j(\mathbf{x}) \leq \|\mathbf{a}^{(l)}\|_\infty \|\text{sign}(\mathbf{a}^{(l)}) - \mathbf{b}^{l}_j\|_1
\end{equation}
where $\text{sign}(x) = 1$ if $x > 0$ and $-1$ otherwise; $\mathbf{b}^{l}_j$ is the binary weight corresponding to class $j$. Because classifier outputs are class-dependent and normalized relative to other classes, confidence scores are not comparable between different classes and do not provide a globally consistent confidence signal. In contrast, the feature norm reflects the model's overall discriminative strength and activation magnitude. Because this property relies on internal activation patterns rather than specific class semantics, the feature norm serves as a \textit{class-agnostic} indicator of confidence~\cite{featurenorm}.  We empirically validate this relationship by visualizing internal activation tendencies. As illustrated in \cref{fig:mnist_norms}(d), these characteristic norms act as a concise summary of the activation tendencies within the hidden layers of the network. For example, for the MNIST dataset, an easily identifiable image, \cref{fig:mnist_norms}(d), shows stronger activations, leading to a high $L1$ norm. In contrast, low-quality images show weaker activations \cref{fig:mnist_norms}(d), leading to a low $L1$ norm. In addition, in \cref{fig:mnist_norms}(c), we plot a scatter plot of the normalized $L_1$ norm (X-axis) and the maximum softmax probability (Y-axis). We make the following observations: (1) Softmax exhibits a saturating tendency, masking fine-grained information; and (2) $L_1$ has a wider dynamic range, which captures subtle variations. Furthermore, even when $L_1$ is low, its corresponding softmax confidence is saturated. In \cref{subsubsec:NCIlabel}, we detail the engineering solution required to adapt this $L_1$ for sample-specific label smoothing.

\section{Method} \label{sec:3.1}

\noindent Label Smoothing (LS) and Margin-based Label Smoothing (MbLS) can be viewed as a cross-entropy-based classification objective augmented with an entropy-penalty regularization term~\cite{mbls}. The trade off between these two components is fixed. We introduce a sample-dependent coefficient $\alpha^{(i)}$ to create a convex objective that adaptively modulates the balance between classification and regularization. This coefficient is calculated from detached features and logits, and is constrained within $[\alpha_{\min}, \alpha_{\max}]$ to prevent the vanishing of the classification signal or the regularization effect during training. 
\textbf{Model Decomposition}. The Neural network classifier can be decomposed as: $f_\omega(\mathbf{x}) = \sigma( F_{\text{head}} \left( F_{\text{backbone}}(\mathbf{x}) \right))$ as shown in \cref{fig:highlvelarchmain}(a)\cite{decoupling}. Here, the feature extractor $F_{\text{backbone}}(\mathbf{x}^{(i)}) = \mathbf{z}^{(i)} \in \mathbb{R}^d$ maps the input to a $d$-dimensional feature space. The classification head $F_{\text{head}}(\mathbf{z}^{(i)}) = \mathbf{s}^{(i)} \in \mathbb{R}^C$ maps the features to the class logits $\mathbf{s}^{(i)} = \{s_{k}^{(i)}\}_{k=1}^{C}$. These logits are normalized via the \textit{softmax function $\sigma(\cdot)$} to produce the predicted probability distribution $\hat{\mathbf{p}}^{(i)} =  \{\hat{p}_{k}^{(i)}\}_{k=1}^{C}$.

\subsubsection{Vanilla Label Smoothing.}

Following the definitions in \cref{premsprobform}, Label Smoothing modifies the original one-hot labels $\mathbf{y^{(i)}} \rightarrow \mathbf{\tilde{y}^{(i)}}$,  with a smoothing parameter $\alpha \in [0, 1]$ such that each element is modified as $\tilde{y}^{(i)}_{k} = (1-\alpha) \cdot y_{k}^{(i)} + \frac{\alpha}{C}$. By applying cross-entropy and rearranging the terms, we get:

\begin{equation} \label{eq:lsdecompsition}
   \mathcal{L}_{\text{LS}}^{i}  = - \sum_{k=1}^{C} \tilde{y}^{(i)}_{k} \log \hat{p}_{k}^{(i)} = (1-\alpha)\underbrace{(-\sum_{k=1}^{C}y^{(i)}_{k} \log {\hat{p}}^{(i)}_{k})}_{\text{Classification Loss}}  +  \alpha\underbrace{(-\sum_{k=1}^{C} \frac{1}{C}\log \hat{p}_{k}^{(i)})}_{\text{Regularization Loss}~E(\mathbf{\hat{p}}^{(i)})},
\end{equation}

The final form of the label-smoothed cross-entropy loss, as shown in \cref{eq:lsdecompsition}, offers a clear interpretation: it is a weighted sum of the cross-entropy-based classification loss with the original one-hot labels and a Kullback-Leibler (KL) divergence term $E(\mathbf{\hat{p}}^{(i)})$ with an additive constant~\cite{mbls}. This KL term encourages the predictive distribution to approach the uniform distribution $\mathbf{u} = [\frac{1}{C}...\frac{1}{C}]$ over all $C$ classes. The smoothing parameter $\alpha$ thus controls the trade-off between fitting the ground truth and distributing the probability mass more broadly, thus regularizing the confidence of the model. To address the limitations of a static parameter, we propose \namels, which extends LS by introducing an adaptive smoothing factor $\alpha^{(i)}$.

\begin{equation} \label{eq:fedlaseq}
   \mathcal{L}_{\namels}^{i}  = (1-\alpha^{(i)})\mathcal{L}^{i}_{\text{CE}} +  \alpha^{(i)}E(\mathbf{\hat{p}}^{(i)})
\end{equation}

\subsubsection{Margin based LS (MbLS).}

\noindent In \textit{MbLS}~\cite{mbls}, the authors characterize the network calibration through the lens of constrained optimization. Within this framework, regularization is viewed as a constraint enforced on \textit{logit gaps}. These gaps are formulated as follows: 
\begin{equation}
%$$.
\mathbf{d}(\mathbf{s}^{(i)}) = \left[\max_{l \in \{1,\cdots,C\}} (s_{l}^{(i)}) - s_{1}^{(i)},\dots,\max_{l \in \{1,\cdots,C\}} (s_{l}^{(i)}) - s_{C}^{(i)}\right]
\end{equation} 

\noindent This perspective reveals that standard calibration methods \cite{ls, ecp, focalloss}, can be reinterpreted as optimization problems subject to a soft equality constraint $\mathbf{d}(\mathbf{s}^{(i)}) = \mathbf{0}$. By maximizing entropy, these methods implicitly drive all logit gaps toward zero. However, such a strict equality constraint is too restrictive as it constantly pushes the model toward a state of maximum uncertainty\cite{mbls}. \textit{MbLS} relaxes this constraint by replacing the equality constraint $\mathbf{d}(\mathbf{s}^{(i)}) = \mathbf{0}$ with a margin based set-constraint: $\mathbf{d}(\mathbf{s}^{(i)}) \leq m \cdot \mathbf{1}_{C}$. As demonstrated in \cite{acls}, this formulation can also theoretically be interpreted as \textit{conditional label smoothing}; specifically, the logit $s_{k}^{(i)}$ is regularized only if the logit gap $\max_{l} (s_{l}^{(i)}) - s_{k}^{(i)}$ exceeds the predefined margin $m$.

\begin{equation} \label{eq:mblsbefroe}
\mathcal{L}_{\text{MbLS}} =  \mathcal{L}^{i}_{\text{CE}} + \lambda \cdot \sum_{k=1}^{C} \max\left(0, \max_{l \in \{1,\cdots,C\}} (s_{l}^{(i)}) - s_{k}^{(i)} -m \right) := \mathcal{L}^{i}_{\text{CE}} +
 \lambda \cdot R(\mathbf{s^{(i)}})
\end{equation}

However, the regularization strength remains \textit{non-adaptive}~\cite{mbls}. A critical weakness identified in \cite{acls} is that the standard additive MbLS penalty is unlikely to regularize the training label of the true class directly. Often, the true logit is the max-logit, therefore, it cancels out: $\max_{l \in \{1,\cdots,C\}} (s_{l}^{(i)}) - s_{k}^{(i)}=0$, which can limit its ability to mitigate overconfidence.  \noindent Our \namembls~approach employs $\alpha^{(i)}$ as a unified balancing parameter in a convex combination, $\mathcal{L}_{\namembls}^{(i)} = (1-\alpha^{(i)})\mathcal{L}^{i}_{\text{CE}} + \alpha^{(i)}R(\mathbf{s}^{(i)})$, while \textit{MbLS}~\cite{mbls} scales the regularization term in isolation without such an interpolation.

\subsection{Adaptive Smoothing Module (ASM) $f(.)$.} \label{MethodsASM}
This plug-in module $f(.)$ modulates the smoothing parameter $\alpha$ per sample, so that for each sample, the smoothing parameter for the corresponding loss is modulated with respect to its features and logits as $\alpha^{(i)}=\alpha \cdot f(\cdot)$ for \namels~or $\alpha^{(i)}=\lambda \cdot f(\cdot)$ for \namembls . $\alpha$ and $\lambda$ refer to the original hyperparameters of LS and MBLS.
\begin{equation}\label{eq:postls1}  
   %\frac{\partial \mathcal{L}_{\text{CE}}}{\partial {z}_{j}} = {p}_{j} - {q}_{j}
    \alpha^{(i)} =  \alpha f( s^{(i)}, z^{(i)}) = \alpha \cdot 2\phi\left(\beta \cdot \psi\left(sg[s^{(i)}]\right) \cdot \gamma\left(sg[z^{(i)}]\right)\right)
\end{equation}
where $\phi(x)  = \frac{1}{1 + e^{-x}}$ is the sigmoid function, $sg[\cdot]$ is the gradient detaching function. $\gamma(z^{(i)}) \in \mathbb{R} $ is a Norm-based Confidence Indicator (NCI). $\psi(s^{(i)}) \in \{-1,1\}$ is the Bidirectional Calibrating Gate (BCG) Module. $\beta$ is a tunable hyperparameter (selected on the validation set) which controls the sensitivity of the smoothing coefficient to changes in the input. Our design choice of sigmoid is influenced by various properties that we wanted to preserve. (1)  \textbf{Boundedness}: The module scales the base hyperparameter $\alpha^{(i)}$ within the range $(0,2\alpha) $(reverting to the baseline method when $\beta=0$). This protects the system from outliers. (2) \textbf{Monotonicity}: The sigmoid function is monotonic. In each operating mode, the smoothing factor is scaled proportionally to the NCI (3) \textbf{Symmetric}: Essentially, our model has two operating modes: positive and negative.  It is determined by the BCG module. Both modes are rotationally symmetric.

\begin{figure}[tb]
  \centering
  \includegraphics[width=0.9\linewidth]{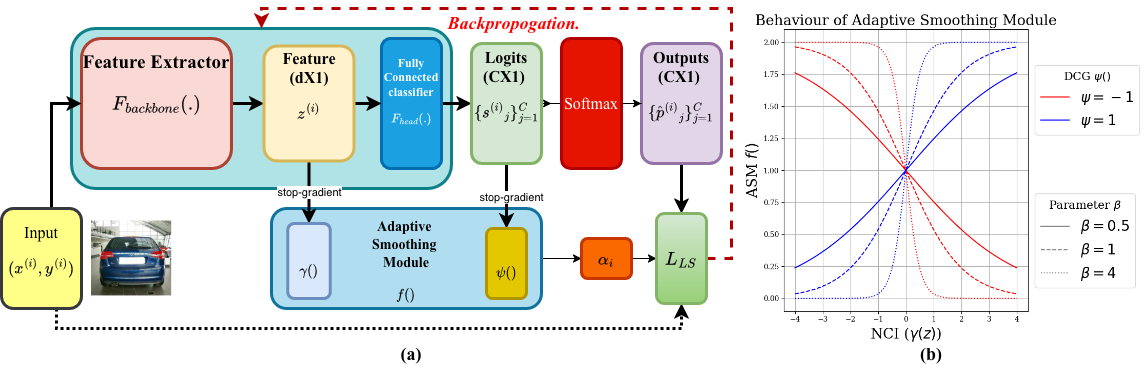}
  \caption{(a) High-Level Architecture and Data flow of \nameplug. A standard feature extractor, such as a ResNet($F_{backbone}(\cdot)$), is connected to a fully connected classifier($F_{head}(\cdot)$). Our proposed Adaptive Smoothing Module (ASM) dynamically adjusts the regularization for each sample by leveraging the detached features/logits extracted by the backbone network. (b) The behavior of Adaptive Smoothing Module (ASM) $f(.)$. The intensity of the scaling is controlled by the \emph{Norm-based Confidence Indicator (NCI)} $\gamma\!\left(z\right)$. This estimates prediction confidence from the norm of the intermediate feature representation $z^{(i)}$. The polarity of scaling, if its outputs are monotonically increasing (blue) /decreasing (red) for the NCI signal is determined by \emph{Bidirectional Calibrating Gate (BCG)} $\psi\left(s\right)$. This compensates for systematic overconfidence or under-confidence for each sample.}
  \label{fig:highlvelarchmain}
\end{figure}

\subsubsection{Bidirectional Calibration Gating Module $\psi(\cdot)$ (BCG Module).}  

The BCG module acts as a discrete binary selector, $\psi(\mathbf{s}^{(i)}) \in \{-1, 1\}$, determining the smoothing regime for each sample:

\begin{equation}
\psi(\mathbf{s}^{(i)}) = 
\begin{cases} 
1, & \text{if \textit{Positive Mode} (Mitigating Overconfidence)} \\
-1, & \text{if \textit{Negative Mode} (Counteracting Underconfidence)}
\end{cases}
\end{equation}

In \textit{Positive Mode}, the relationship between the norm-based confidence indicator $\gamma(\cdot)$  and the smoothing parameter $\alpha^{(i)}$ is monotonically increasing ($\frac{d\alpha^{(i)}}{d\gamma^{(i)}}~>~0$; \cref{fig:highlvelarchmain} (b) (Blue)), provides an ``up-regulation'' that prevents pathological overconfidence. Conversely, in \textit{Negative Mode}, the relationship is monotonically decreasing ($\frac{d\alpha^{(i)}}{d\gamma^{(i)}} < 0$; \cref{fig:highlvelarchmain} (b) (Red)), allowing for a ``down-regulation'' that boosts the model's confidence where it is otherwise lacking.

To determine the optimal regime, we employ a lightweight gating network $F_{gate}: \mathbb{R}^C \to \mathbb{R}^2$ that processes logits detached from the gradient graph $\text{sg}[\mathbf{s}^{(i)}]$. To maintain end-to-end differentiability despite the discrete nature of $\psi(\cdot)$, we utilize a \textbf{Straight-Through Estimator (STE)}\cite{ste}. The gating logic is formulated as follows:

\begin{equation}
\psi(\mathbf{s}^{(i)}) = \text{sgn} \left( \mathbf{w}_{\text{gate}}^\top \cdot \sigma \left( \mathcal{F}_{gate} ( \text{sg}[\mathbf{s}^{(i)}] ) \right) \right), \quad \text{with } \mathbf{w}_{\text{gate}} = [1, -1]^\top
\end{equation}

where $\text{sgn}(x) = \left\{ 1 \text{ if } x > 0, -1 \text{ otherwise} \right\}$, $\mathcal{F}_{gate}$ represents a Multi-Layer Perceptron (MLP), and $\sigma$ denotes the softmax function. This formulation effectively maps the predicted state to a bipolar directional signal.

\subsubsection{Norm-based Confidence Indicator $\gamma(\cdot)$ (NCI) } \label{subsubsec:NCIlabel}

To quantify sample confidence, we define a \textit{Norm-based Confidence Indicator (NCI)}, denoted as $\gamma(\cdot)$. Instead of using the raw feature norms, we normalize it with Exponential Moving Averages. The reasons for this are as follows: (1) Without mean-variance normalization, fluctuations in norm variance cause the modulation signal to frequently enter saturation regions of the sigmoid function in the ASM module, leading to "dead" adaptation. ; (2) By using an exponentially moving baseline, normalization ensures that the smoothing correction for each sample is approximately determined by its relative position within the evolving population of training points, rather than within the current batch. The NCI for a sample $i$ is calculated as:
\begin{equation} 
\gamma(z^{(i)}) = \frac{ \|sg[z^{(i)}]\|_1 - \mu^{k}_z}{\sigma^{k}_z}
\end{equation} 

where $\| \cdot \|_1$ represents the $L_1$ norm and $sg[\cdot]$ denotes the stop-gradient operation \cite{ste}. We use $sg[\cdot]$ to ensure that the normalization statistics serve purely as a measurement tool and do not influence the underlying feature optimization during backpropagation. The mean ($\mu^{k}_z$) and the standard deviation ($\sigma^{k}_z$) are tracked using EMA:
\begin{align} 
\mu^{k}_z &= \theta \widehat{\mu}^{k}_{z} + (1-\theta)\mu^{k-1}_{z} \\
\sigma^{k}_z &= \theta\widehat{\sigma}^{k}_{z} + (1-\theta)\sigma^{k-1}_{z}
\end{align} 
In these updates, $\widehat{\mu}^{k}_{z}$ and $\widehat{\sigma}^{k}_{z}$ represent the statistics of the current batch $k$, while $\mu$ and $\sigma$ represent the global moving statistics. The momentum parameter $\theta$ controls the influence of the current batch on the moving averages: a higher $\theta \in (0,1)$ allows the averages to adapt quickly to recent statistics, while a lower $\theta$ provides greater stability by weighting past averages more heavily. Furthermore, $\gamma(\cdot)$ can be instantiated using the maximum softmax probability/maximum logit (This is ablated).

\section{Experiments}
To thoroughly validate the effectiveness of our proposed method, we conducted extensive experiments in multiple classification regimes. For the standard classification, the results are summarized in \cref{tab:sota}. We further evaluate our approach on high-resolution \textit{fine-grained classification}, a challenging task that requires distinguishing between subordinate categories within a single entry-level class (\eg, specific bird species or aircraft variants). Unlike standard recognition, fine-grained tasks are characterized by subtle differences between classes and high intra-class variance; performance on these high-resolution benchmarks is reported in \cref{tab:finegrained}. 

\noindent \textbf{Datasets.} 
For standard classification, we utilize the widely-used CIFAR-10, CIFAR-100~\cite{krizhevsky2009learning}, and Tiny-ImageNet~\cite{deng2009imagenet} benchmarks. For fine-grained visual classification (FGVC), we employ the CUB-200-2011~\cite{WahCUB_200_2011} and FGVC-Aircraft~\cite{Maji2013FineGrained}.

\noindent \textbf{Architectures.} 
Following \cite{acls}, for standard image classification (CIFAR/Tiny-ImageNet), we train \textit{ResNet-50} and \textit{ResNet-101}~\cite{resnet} architectures from scratch. For fine-grained tasks, we use the ImageNet-1K pre-trained \textit{ResNet-101} (following \cite{mbls}) and, beyond prior work, the ImageNet-1K pre-trained \textit{ViT-B/16}~\cite{ViT_ICLR2021dosovitskiy}.

\noindent \textbf{Experimental Setup.} 
All experiments were conducted on a single NVIDIA A100 (40GB) GPU running Linux. To ensure a strictly fair comparison, our training protocol, including the optimizer, scheduler, and data augmentations, perfectly replicate the standardized setup established by ACLS\cite{acls}. Detailed training protocols, image pre-processing steps, and augmentation strategies are elaborated in the Appendix. 

\noindent \textbf{Metrics.} Classification performance is quantified using \textit{Top-1 Accuracy}. Model calibration is assessed primarily via the \textit{Expected Calibration Error (ECE)} and \textit{the Adaptive Expected Calibration Error (AECE)}~\cite{mbls,ece,eace}. Following \cite{acls}, all calibration metrics are calculated using the $M=15$ bins. AECE is more robust than ECE due to its adaptive binning strategy~\cite{eace}.

\noindent \textbf{Implementation details of the compared baselines.}
As our approach is a training-time calibration method based on entropy regularization, we compare it against a broad spectrum of established training-time techniques with implicit/explicit entropy penalties. We include several contemporary methods such as MMCE~\cite{mmca}, MDCA~\cite{MDCA}, CPC~\cite{CPC}, and CRL~\cite{CRLcheck}.

We also consider the original Focal Loss (FL)~\cite{focalloss} and its extensions: Sample Dependent Focal Loss (FLSD)~\cite{calibfocal}, AdaFocal~\cite{adafocal}, and Dual Focal Loss (DFL)~\cite{dfl}. The hyperparameters for these methods align with those specified in their respective original publications. Furthermore, we evaluated against standard Label Smoothing (LS) with a fixed factor $\alpha=0.05$ (Exception of $\alpha=0.1$ for CIFAR100), as well as Margin-Based Label Smoothing (MbLS)~\cite{mbls} and its extension, ACLS~\cite{acls}. For these, we follow the original configurations: $m=6, \lambda=0.1$ for CIFAR-10/100 and $m=10, \lambda=0.05$ for Tiny-ImageNet, CUB-2011, and FGVC-Aircraft. Within this comparison, our primary state-of-the-art (SOTA) baselines are \textit{AdaFocal\cite{adafocal}, Dual Focal Loss\cite{dfl}, and ACLS\cite{acls}}. We include Temperature Scaling~\cite{guo2017calibration} as a representative post-hoc baseline, but we do not compare against other post-hoc methods because our approach is an in-training objective that actively shapes the model during learning, rather than an adjustment applied to a fixed, trained model.

\noindent \textbf{Implementation details of the proposed method.} 
We integrate our proposed plug-in, hereafter referred to as \nameplug, into two distinct baseline losses: standard Label Smoothing (LS) and Margin-based Label Smoothing (MbLS)~\cite{mbls}. This results in two configurations: \namels~and \namembls. For the \namels~and \namembls~configurations, all original LS and MbLS hyper-parameters are kept unchanged for each dataset to ensure a controlled evaluation. To compute the required feature norms, we utilize the pre-classifier feature vectors for ResNet architectures and the global class token (CLS) for Vision Transformers. Our method introduces a dataset-specific hyperparameter, $\beta$, and a batch-size-specific momentum smoothing parameter, $\theta$. The optimal values for these were determined through the validation set. Specifically, we found $\beta=4.0$ to be effective for the standalone \namels~baselines in all datasets, while $\beta=0.5$ performed optimally for the \namembls~configuration in CIFAR-10 and Tiny-ImageNet, and $\beta=1$ performed optimally for the \namembls~configuration in CIFAR-100. The momentum parameter $\theta$ was set to 0.95 for CIFAR-10 and 0.5 for Tiny-ImageNet and CIFAR-100. For the standard classification benchmarks, the hyperparameters were tuned directly in the validation sets. However, since fine-grained datasets typically lack a formal validation split and have a smaller training dataset size, we followed the protocol established in~\cite{mbls} and adopted the optimal hyper-parameters identified for Tiny-ImageNet for all experiments involving CUB-2011 and FGVC-Aircraft.

\begin{table}[tb]
  \caption{Results on the Standard Image Recognition/Calibration test splits. We report the Top-1 Accuracy, ECE, AECE. Calibration metrics calculated with 15 bins. Results follow the unified setup of~\cite{acls}; values are from~\cite{acls}, except rows marked $\dagger$, which we run as means over three fixed seeds. The best-performing result in each category is indicated in \textbf{bold}, second-best is \underline{underlined}. The final three columns show the average accuracy across datasets/backbones (expanded in Supplemental) and the mean rank of each method for ECE and AECE across all tested scenarios}
  \label{tab:sota}
  \centering

\resizebox{1\textwidth}{!}{
    \begin{tabular}{@{} l cc cc cc cc cc cc ccc @{}}
        \toprule
        & \multicolumn{4}{c}{Tiny-ImageNet\cite{deng2009imagenet}} & \multicolumn{4}{c}{CIFAR10\cite{krizhevsky2009learning}} & \multicolumn{4}{c}{CIFAR100\cite{krizhevsky2009learning}} & \multicolumn{3}{c}{Average (Dataset)} \\
        \cmidrule(lr){2-5} \cmidrule(lr){6-9} \cmidrule(lr){10-13} \cmidrule(l){14-16}
        Arch. & \multicolumn{2}{c}{ResNet50} & \multicolumn{2}{c}{ResNet101} & \multicolumn{2}{c}{ResNet50} & \multicolumn{2}{c}{ResNet101} & \multicolumn{2}{c}{ResNet50} & \multicolumn{2}{c}{ResNet101} & Value & Rank & Rank \\
        Metrics & {\scriptsize ECE$\downarrow$} & {\scriptsize AECE$\downarrow$} & {\scriptsize ECE$\downarrow$} & {\scriptsize AECE$\downarrow$} & {\scriptsize ECE$\downarrow$} & {\scriptsize AECE$\downarrow$} & {\scriptsize ECE$\downarrow$} & {\scriptsize AECE$\downarrow$} & {\scriptsize ECE$\downarrow$} & {\scriptsize AECE$\downarrow$} & {\scriptsize ECE$\downarrow$} & {\scriptsize AECE$\downarrow$} & {\scriptsize Acc.$\uparrow$} & {\scriptsize ECE$\downarrow$} & {\scriptsize AECE$\downarrow$} \\
        \midrule

 \multicolumn{16}{@{}l}{\textit{Training-time Methods}} \\
   CE\cite{ce}  & 3.73 & 3.69 & 4.97 & 4.97 & 5.85 & 5.84 & 5.74 & 5.73 & 13.59 & 13.54 & 12.94 & 12.94 & 78.74 & 14.00 & 14.00 \\
   MMCE\cite{mmca}  & 5.15 & 5.12 & 4.88 & 4.88 & 3.10 & 3.10 & 3.61 & 3.61 & 12.72 & 12.71 & 13.43 & 13.42  & 79.37 & 11.50 & 11.00 \\ 
   ECP\cite{ecp}  & 4.00 & 3.92 & 4.68 & 4.66 & 3.01 & 2.99 & 5.41 & 5.40 & 12.29 & 12.28 & 13.43 & 13.42 & 78.63 & 11.50 & 11.00 \\
   MDCA\cite{MDCA}  & 2.77 & 2.61 & 6.06 & 6.06 & 6.86 & 6.73 & 6.88 & 7.23 & 12.68 & 12.65 & 13.61 & 13.59 & 79.40 & 14.17 & 14.17 \\
   CPC\cite{CPC} & 3.12 & 3.05 & 3.90 & 4.00 & 3.91 & 3.91 & 3.78 & 3.75 & 13.29 & 13.28 & 13.32 & 13.28 & 79.52 & 12.00 & 11.50 \\
   CRL\cite{CRLcheck}  & 1.65 & 1.52 & 3.57 & 3.56 & 3.14 & 3.11 & 3.74 & 3.73 & 6.30 & 6.26 & 7.29 & 7.14  & 79.39 & 8.50 & 7.50 \\
   FLSD-53\cite{calibfocal}  & 2.91 & 2.95 & 4.91 & 4.91 & 3.84 & 3.60 & 4.58 & 4.57 & \underline{4.23} & 4.21 & 5.33 & 5.26  & 78.36 & 8.83 & 8.67\\
   FL\cite{focalloss} & 2.96 & 3.12 & 2.55 & 2.44 & 3.90 & 3.86 & 4.60 & 4.58 & 4.81 & 4.79 & 5.13 & 5.14 & 78.26 & 8.67 & 8.83 \\
   DFL\cite{dfl} $\dagger$ & 6.50 & 6.50 & 4.78 & 4.77 & 2.46 & \underline{2.39} & 2.12 & \underline{2.15} & 4.55 & 4.38 & 5.05 & 5.10 & 78.89 & 7.50 & 6.33 \\
         AdaFocal\cite{adafocal} $\dagger$ & 5.82 & 5.80 & 5.43 & 5.42 & 4.9 & 4.81 & 3.52 & 3.29 & 4.39 & 4.39 & 5.31 & 5.22 & 78.77 & 10.00 & 9.83 \\

      LS\cite{ls} $\dagger$ & 2.56 & 2.58 & 2.04 & 1.96 & 3.59 & 4.09 & 3.07 & 4.18 & 5.42 & 5.72 & 6.08 & 8.87 & \underline{79.69} & 7.00 & 9.00 \\

   \namels (Ours) $\dagger$ & 1.77 & 1.65 & 2.56 & 2.39 & 1.29 & \textbf{2.30} & 1.56 & \textbf{1.92} & 4.29 & \underline{4.20} & 7.48 & 7.27 & 79.62 & 5.50 & \underline{4.17}\\  
   
   MbLS\cite{mbls} $\dagger$ & 1.45 & 1.47 & 1.81 & \underline{1.58} & \underline{1.21} & 3.34 & \underline{1.38} & 3.25 & 5.00 & 5.52 & 6.10 & 10.07 & 79.66 & \underline{4.33} & 6.00 \\  
    ACLS\cite{acls} $\dagger$ & \underline{1.37} & \underline{1.36} & \underline{1.66} & 1.65 & 2.32 & 2.99 & 2.22 & 2.88 & 7.05 & 7.32 & 7.09 & 7.11  & \bf{79.78} & 5.50 & 4.83 \\ 
   \namembls (Ours) $\dagger$ & \textbf{1.22} & \textbf{1.11} & \textbf{1.55} & \textbf{1.39} & \textbf{1.12} & 3.20 & \textbf{1.27} & 3.58 & 4.42  & 4.48 & \underline{4.51} & \underline{4.42} & 79.58 & \textbf{1.83} & \textbf{3.83} \\ \hline
 \multicolumn{16}{@{}l}{\textit{Post-hoc Methods}} \\
         CE+ TS  \cite{guo2017calibration}     & 1.63 & 1.52 & 2.08 & 2.03 & 3.68 & 3.67 & 3.62 & 3.62 & \bf{2.93} & \bf{2.86} & \bf{3.06} & \bf{2.80} & 78.74 & 5.00  & 4.83 \\ \hline

    \end{tabular}} 
\end{table}

\begin{table}[tb]
  \caption{Results on the Fine-Grained High-Resolution Classification test splits\cite{WahCUB_200_2011}. We report the Top-1 Accuracy, ECE, and AECE. Calibration metrics are calculated with 15 bins. The best-performing result in each category is indicated in \textbf{bold}, second-best is \underline{underlined}. Following the same evaluation protocol in \cite{mbls}, all relevant hyper-parameters were selected on a held-out proxy dataset (Tiny-ImageNet) and kept fixed across all benchmarks. All methods use a single fixed seed due to compute cost.}
  \label{tab:finegrained}
  \centering

\resizebox{1\textwidth}{!}{
\begin{tabular}{lccccccccccccccc}
\toprule
\textbf{Arch.} & \multicolumn{6}{c}{FGVC-Aircraft\cite{Maji2013FineGrained}} & \multicolumn{6}{c}{CUB-2011-Birds\cite{WahCUB_200_2011}} & \multicolumn{3}{c}{Average} \\
\cmidrule(r){2-7} \cmidrule(l){8-13}
\textbf{Metrics} & \multicolumn{3}{c}{ResNet-101} & \multicolumn{3}{c}{ViT-B/16 @384} & \multicolumn{3}{c}{ResNet-101} & \multicolumn{3}{c}{ViT-B/16 @384} & Value & Rank & Rank \\
\cmidrule(r){2-4} \cmidrule(lr){5-7} \cmidrule(lr){8-10} \cmidrule(l){11-13}
& Acc.$\uparrow$ & ECE$\downarrow$ & AECE$\downarrow$ & Acc.$\uparrow$ & ECE$\downarrow$ & AECE$\downarrow$ & Acc.$\uparrow$ & ECE$\downarrow$ & AECE$\downarrow$ & Acc.$\uparrow$ & ECE$\downarrow$ & AECE$\downarrow$ & Acc.$\uparrow$ & ECE$\downarrow$ & AECE$\downarrow$ \\
\midrule
CE\cite{ce}    & 87.42 & 4.79 & 4.72 & 68.01 & 25.62 & 25.61 & 70.83 & 12.12 & 12.12 & 65.62 & 26.90 & 26.89 & 72.97 & 8.75& 8.75\\
MMCE\cite{mmca}  & 88.44 & 3.88 & 3.87 & 67.23 & 25.35 & 25.34 & 72.55 & 8.82     & 8.71 & 76.18 & 17.27 & 17.26 & 76.10 & 7.75& 7.75\\ 
%FL\cite{focalloss}    & 87.93 & 4.05 & 3.42 & 67.74 & 11.59 & 9.89  & 71.26 & 2.19 & 1.99 & 73.90 & 14.58 & 14.58 \\

DFL\cite{dfl}   & 86.79  & 2.80 & \underline{2.39} & 64.41& 24.67 & 24.67 & 71.34 & \textbf{2.75} & \textbf{2.74} & 74.93  & 15.97 & 15.91 & 74.36 & 4.75& 4.25\\

AdaFocal\cite{adafocal}   & 88.02 & 9.31 & 9.08 & 67.71 & \underline{17.76} & \underline{17.75} & 71.24 & 7.63 & 7.63 & 74.64 & 11.90 & 11.90 & 75.40 & 5.50 & 5.50\\

ACLS\cite{acls}  & 87.93 & 3.51 & 3.67 & 73.02 & 22.40 & 22.40 & \textbf{72.93} & 7.11 & 7.06 & \textbf{83.58} & 11.82 & 11.81 & \bf{79.36} & 5.00 & 5.00\\ 

LS\cite{ls}    & 88.39 & 3.01 & 2.61 & 69.21& 20.53 & 20.44 & 71.72 & \underline{3.51} & 3.65 & 82.74 & \underline{5.36} & \underline{6.43} & 78.01 & \underline{3.00} & \underline{3.00}\\
\namels (Ours) & \underline{88.50} & \bf{1.53} & \bf{1.44} & \underline{73.17} & \bf{12.38} & \bf{12.11} & 72.67 & 3.62 & \underline{3.39} & 82.33 & \bf{2.85} & \bf{2.74}  & 79.16 & \textbf{1.50} & \textbf{1.25}\\
MbLS\cite{mbls}  & 88.35 & 2.47 & 2.80 & \textbf{73.29} & 21.89 & 21.86 & 72.45 & 5.62 & 5.38 & 82.49 & 12.40 & 12.42 & 79.14 & 4.50 & 5.00\\

\namembls (Ours) &  \textbf{88.71} & \underline{2.16} & 2.55 & 72.00 & 23.99 & 23.98 & \underline{72.73} & 4.41 & 4.20  & \underline{83.32} & 12.03 & 12.03 & \underline{79.19} & 4.25 & 4.50\\ 
\bottomrule
   \end{tabular}}
\end{table}

\section{Results, Discussion and Ablation Studies}
%\footnote{Extensive ablation study is in the supplemental}

\subsubsection{\nameplug~is an Effective Plug-in for LS and MbLS.} To analyze the results, we calculate the mean rank for each of the metrics (AECE, ECE) across all dataset-backbone combinations for each method for both standard classification datasets (\cref{tab:sota}) and fine-grained high-resolution classification datasets \cref{tab:finegrained}. These ranks, displayed in the final columns of \cref{tab:sota} and \cref{tab:finegrained}, provide a holistic view of performance beyond individual data set wins. In standard benchmarks, \namembls~has the best (lowest) rank among all compared methods for both calibration metrics. In particular, \namembls~shows a significant ranking boost over its base method, MbLS, suggesting that our proposed modifications effectively bridge the calibration gap found in margin-based losses. Similarly, \namels~consistently outperforms label smoothing (LS), maintaining a lower mean rank. The advantages of our approach are even more pronounced in the fine-grained setting (\cref{tab:finegrained}). The winner across settings depends entirely on whether a specific task is fundamentally better suited
for a LS or MbLS. Based on the results in \cref{tab:sota},\cref{tab:finegrained}, as a rule of thumb, we recommend \namembls~for standard classification and \namels~for fine-grained classification.

\subsubsection{Balance of Accuracy and Calibration.} A perennial challenge in model calibration is the potential degradation of Top-1 accuracy \cite{guo2017calibration}. To evaluate this trade-off, we calculate the mean Top-1 accuracy across all data set-backbone combinations for standard (\cref{tab:sota}) and fine-grained high-resolution datasets (\cref{tab:finegrained}) (Extended results are available in the Supplementary Material). Standard Classification: As shown in \cref{tab:sota}, the addition of \nameplug~does not compromise predictive performance. Both variants \namels~(79.62\% vs. 79.69\%) and \namembls~(79.58\% vs. 79.66\%) maintain high accuracies competitive to their baselines (LS, MbLS). Notably, our methods outperforms most existing calibration techniques; for instance, \nameplug~variants (79.58\%–79.62\%) consistently exceed the accuracy range of Focal Loss-based methods \cite{adafocal, dfl, focalloss}, which typically fall between 78.26\% and 78.89\%.  This trend is even more pronounced in the fine-grained setting (\cref{tab:finegrained}). Our methods (ranging from 79.16\% to 79.19\%) significantly outperform Focal Loss variants (74.36\%–75.40\%), while showing a slight improvement over the base LS (78.01\%) and MbLS (79.14\%) models. These results demonstrate that \nameplug~not only enhances calibration but also serves as a robust objective for maintaining, or even slightly improving, Top-1 accuracy in complex classification tasks.

\subsubsection{\nameplug~Improves OOD Detection.} \cref{tab:ood_transposed_v2} compares Out-of-Distribution (OOD)\cite{ood} detection performance (AUROC \%) across standard CIFAR benchmarks. We observe that both static LS and MbLS consistently degrade OOD discriminative power compared to the Cross-Entropy (CE) baseline across all training-testing configurations, indicating a clear trade-off between calibration and OOD separation. In particular, the integration of our \nameplug~plug-in consistently narrows this performance gap.

\begin{table}[tb]
  \caption{\textbf{OOD detection performance (AUROC \%) } Comparison across CIFAR benchmarks for standard and augmented objectives. Our \nameplug~plug-in increases the OOD discriminative power of the base objective, closing the gap with Cross-entropy. Additional baselines are provided for reference. Uses trained models from \cref{tab:sota}.
  }
  \label{tab:ood_transposed_v2}
  \centering

\resizebox{1\textwidth}{!}{\begin{tabular}{@{} lll ccccccccc @{}}
\toprule
\multicolumn{2}{c}{\textbf{Datasets}} & & \multicolumn{5}{c}{\textbf{Baselines}} & \multicolumn{2}{c}{\textbf{LS-based}} & \multicolumn{2}{c}{\textbf{MbLS-based}} \\
\cmidrule(lr){4-8} \cmidrule(lr){9-10} \cmidrule(lr){9-10}
\textbf{Training} & \textbf{Testing} & \textbf{Metric} & CE & TS & DFL & AdaFocal & ACLS  & LS & \textbf{\namels} & MbLS & \textbf{\namembls} \\
\midrule

\textbf{C10}   & SVHN & AUROC$\uparrow$ & 91.37 & 92.20 & 93.87 & 96.37 & 89.14 & 74.23 & \textbf{85.04} & 89.90 & \textbf{91.40} \\
              & C100 & AUROC$\uparrow$ & 87.82 & 88.00 & 86.43 & 87.30 & 84.25 & 75.20 & \textbf{77.20} & 77.20 & \textbf{84.15} \\
\midrule

\textbf{C100}   & SVHN & AUROC$\uparrow$ & 79.47 & 79.77 & 81.85 & 83.57 & 75.09 & 72.92 & \textbf{84.33} & 73.88 & \textbf{78.60} \\
              & C10  & AUROC$\uparrow$ & 74.82 & 74.84 & 77.60 & 76.87 & 71.76 & 72.33 & \textbf{73.71} & 73.19 & \textbf{73.72} \\
\bottomrule
\end{tabular}}
\end{table}

% \subsubsection{Ablation Study}
% We perform ablation studies on BCG and further ablations are in the Supplementary.

\subsubsection{BCG is Stable}
\begin{figure*}[t]
    \centering
    % --- (a) Confidence-state distribution ---
    \begin{subfigure}[b]{0.45\textwidth}
        \centering
        \includegraphics[width=\linewidth]{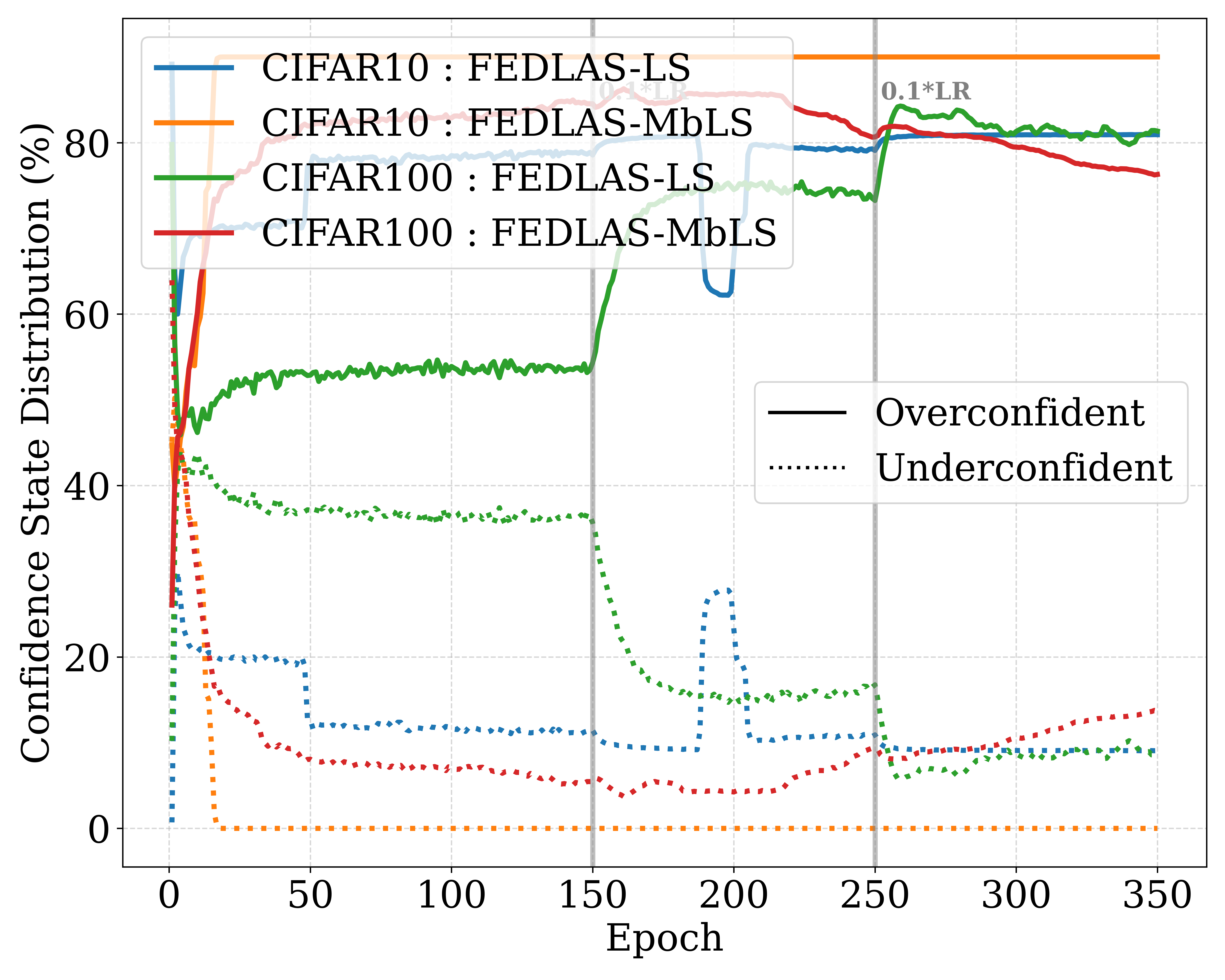}
        \caption{Confidence-state distribution}
        \label{fig:confidencestate}
    \end{subfigure}
    \hfill
    % --- (b) Flip rate ---
    \begin{subfigure}[b]{0.45\textwidth}
        \centering
        \includegraphics[width=\linewidth]{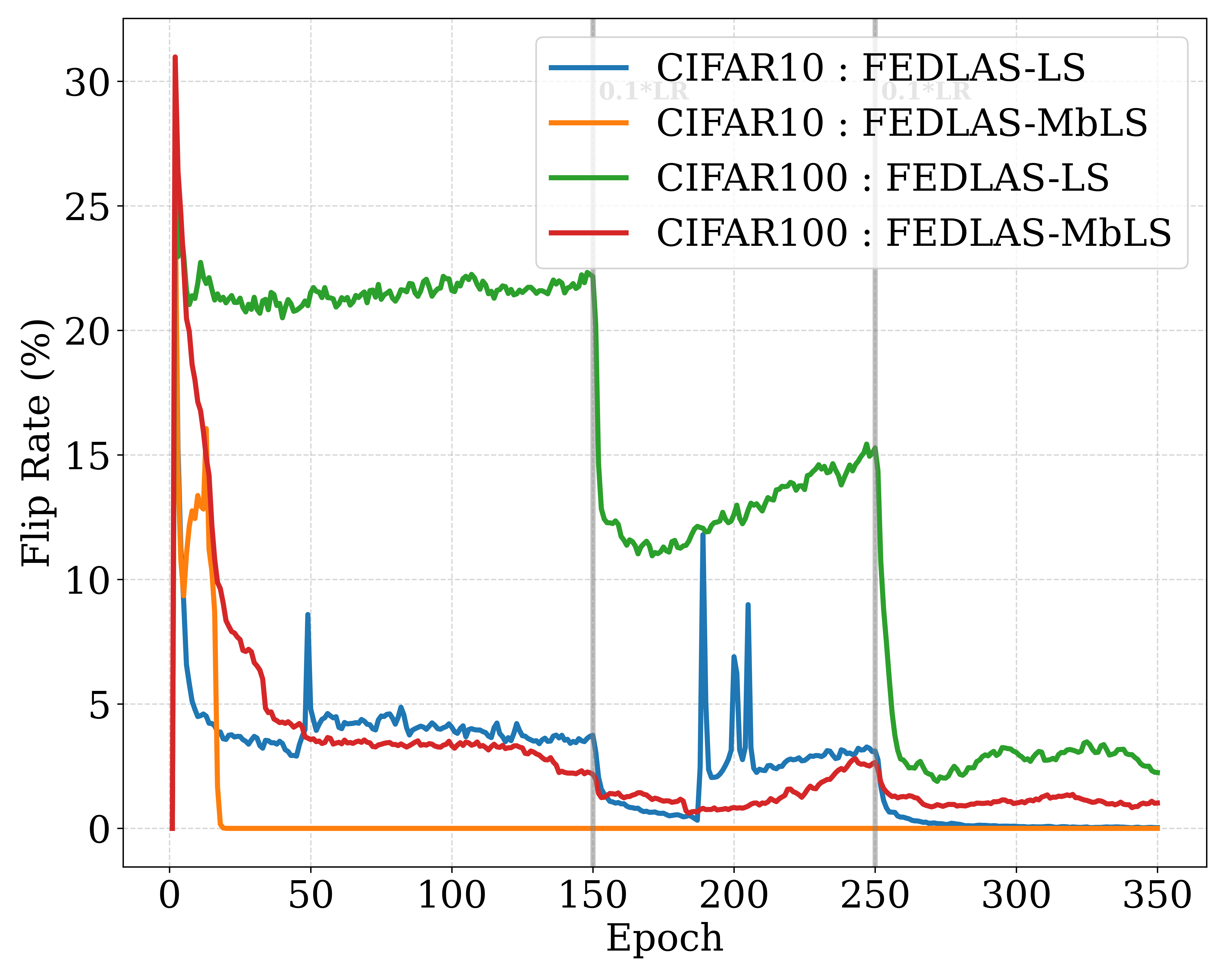}
        \caption{Flip Rate (FR)}
        \label{fig:fliprate}
    \end{subfigure}
    \caption{\textbf{Stability of the BCG module.}
    (a) Proportion of samples classified as overconfident vs.\ under-confident across training epochs. The proportions stay steady, shifting only at learning-rate scheduler steps (epochs $150$, $250$).
    (b) Flip Rate (FR), the average per-epoch rate at which samples switch state. The steady decline shows the gating converges and stabilizes over training.}
    \label{fig:bcg_stability}
\end{figure*}

We track each sample's state by its index across training epochs. Using the BCG module's output, we record whether a sample is in an overconfident or under-confident state. In \cref{fig:confidencestate}, we plot the proportion of samples in each state across the entire dataset per epoch. The proportions remain steady, with shifts occurring mainly when the learning-rate scheduler triggers (epochs $150$ and $250$), indicating stability. In \cref{fig:fliprate}, following the same methodology, we introduce the Flip Rate (FR), which measures how frequently a sample switches state between consecutive epochs on average. The FR decreases steadily, indicating that the gating stabilizes and converges.

\subsubsection{BCG Rectifies Both Underconfidence and Overconfidence}
We decompose ECE into overconfident ECE (O-ECE) and underconfident ECE (U-ECE) and report it in \cref{tab:over_under_ece}. We compare our \nameplug~variants against their base losses (LS, MbLS) and the Cross-Entropy (CE) baseline on standard benchmarks. The results reveal a clear trade-off. CE is strongly overconfident, so while its U-ECE is near zero (it rarely makes underconfident predictions), its O-ECE is large. The base losses LS and MbLS reduce O-ECE but over-correct, shifting many predictions into underconfidence and substantially raising U-ECE. Our \nameplug~variants reduce O-ECE while keeping U-ECE low, achieving the best balance across all three benchmarks. This supports our central claim: bidirectional gating corrects miscalibration in \emph{both} directions rather than trading one failure mode for the other.

\begin{table}[tb]
\scriptsize
  \caption{\textbf{Decomposed calibration error (ECE \%)} Comparison across standard benchmarks on the decomposed ECE (overconfident (O-ECE) and underconfident (U-ECE)). Our \nameplug~plug-in reduces calibration error in \emph{both} directions, whereas the base objectives lower O-ECE only by inflating U-ECE.
  }
  \label{tab:over_under_ece}
  \centering
\begin{tabular}{@{} ll c cc cc @{}}
\toprule
\textbf{Dataset} & \textbf{Metric} & CE & LS & \textbf{\namels} & MbLS & \textbf{\namembls} \\
\midrule
\textbf{C10}   & O-ECE$\downarrow$ & 5.85  & 1.16 & \textbf{0.89} & 1.09 & \textbf{1.11} \\
               & U-ECE$\downarrow$ & 0.00  & 2.45 & \textbf{0.40} & 0.12 & \textbf{0.02} \\
\midrule
\textbf{C100}  & O-ECE$\downarrow$ & 13.59 & 2.73 & \textbf{3.70} & 2.62 & \textbf{2.27} \\
               & U-ECE$\downarrow$ & 0.00  & 2.69 & \textbf{0.59} & 2.38 & \textbf{2.15} \\
\midrule
\textbf{Tiny}  & O-ECE$\downarrow$ & 3.67  & 0.19 & \textbf{1.30} & 0.64 & \textbf{0.46} \\
               & U-ECE$\downarrow$ & 0.05  & 2.37 & \textbf{0.48} & 0.81 & \textbf{0.76} \\
\bottomrule
\end{tabular}
\end{table}

\section{Conclusions, Limitations and Future Work}

\nameplug~is an adaptive label smoothing framework that improves calibration by dynamically adjusting smoothing. By addressing both under- and over-confidence, it consistently outperforms state-of-the-art methods. Furthermore, the introduction of a class-agnostic confidence indicator enables future work in unsupervised, self-supervised, semi-supervised and noisy-label calibration. Limitations-wise, the NCI relies on the L1 feature norm as a confidence proxy, which inherits the assumptions of the underlying theory.

\section*{Acknowledgment}
NB acknowledges Melbourne Graduate Research Scholarship. We would like to thank Chathura Jayasankha, Asela Hevapathige, and Zhaoyuan Qiu for providing valuable feedback. This research was supported by The University of Melbourne’s Research Computing Services.

\bibliographystyle{splncs04}
\bibliography{main}

@String(CVPR  = {IEEE Conf. Comput. Vis. Pattern Recog.})

@String(ICCV  = {Int. Conf. Comput. Vis.})

@String(ICML  = {Int. Conf. Mach. Learn.})

@String(AAAI  = {AAAI})

@String(CVPR  = {CVPR})

@String(ICCV  = {ICCV})

@String(ICML  = {ICML})

@inproceedings{featurenorm,
  title={{Understanding the Feature Norm for Out-of-Distribution Detection}},
  author={Jaehyeok Park and Jin Chye Lih Chai and Jongyoon Yoon and Albert B. J. Teoh},
  booktitle={{Proceedings - 2023 IEEE/CVF International Conference on Computer Vision, ICCV 2023}},
  pages={1557-1567},
  year={2023},
  organization={Institute of Electrical and Electronics Engineers Inc.}
}

@techreport{WahCUB_200_2011,
  author       = {Catherine Wah and Steve Branson and Peter Welinder and Pietro Perona and Serge Belongie},
  title        = {The Caltech-UCSD Birds-200-2011 Dataset},
  institution  = {California Institute of Technology},
  year         = {2011},
  number       = {CNS-TR-2011-001},
  type         = {Technical Report}
}

@techreport{krizhevsky2009learning,
  title={Learning multiple layers of features from tiny images},
  author={Krizhevsky, Alex and Hinton, Geoffrey},
  institution={University of Toronto},
  year={2009},
  url={https://www.cs.toronto.edu/~kriz/learning-features-2009-TR.pdf}
}

@inproceedings{deng2009imagenet,
  title={{ImageNet}: A large-scale hierarchical image database},
  author={Deng, Jia and Dong, Wei and Socher, Richard and Li, Li-Jia and Li, Kai and Fei-Fei, Li},
  booktitle={2009 IEEE Conference on Computer Vision and Pattern Recognition},
  pages={248--255},
  year={2009},
  organization={IEEE},
  doi={10.1109/CVPR.2009.5206848}
}

@inproceedings{guo2017calibration,
author = {Guo, Chuan and Pleiss, Geoff and Sun, Yu and Weinberger, Kilian Q.},
title = {On calibration of modern neural networks},
year = {2017},
publisher = {JMLR.org},
booktitle = {Proceedings of the 34th International Conference on Machine Learning - Volume 70},
pages = {1321–1330},
numpages = {10},
location = {Sydney, NSW, Australia},
series = {ICML'17}
}

@INPROCEEDINGS{Tomani2021Posthoc,
  author={Tomani, Christian and Gruber, Sebastian and Erdem, Muhammed Ebrar and Cremers, Daniel and Buettner, Florian},
  booktitle={2021 IEEE/CVF Conference on Computer Vision and Pattern Recognition (CVPR)}, 
  title={Post-hoc Uncertainty Calibration for Domain Drift Scenarios}, 
  year={2021},
  volume={},
  number={},
  pages={10119-10127},
  doi={10.1109/CVPR46437.2021.00999}}

@inproceedings{zhang2020mix,
author = {Zhang, Jize and Kailkhura, Bhavya and Han, T. Yong-Jin},
title = {Mix-n-Match: ensemble and compositional methods for uncertainty calibration in deep learning},
year = {2020},
publisher = {JMLR.org},
booktitle = {Proceedings of the 37th International Conference on Machine Learning},
articleno = {1031},
numpages = {12},
series = {ICML'20}
}

@inproceedings{largemargin,
  author    = {Platt, John},
  title     = {Probabilistic Outputs for Support Vector Machines and Comparisons to Regularized Likelihood Methods},
  booktitle = {Advances in Large Margin Classifiers},
  year      = {1999},
  pages     = {61--74},
  publisher = {MIT Press}
}

@inproceedings{ovadia2019can,
  title={Can You Trust Your Model's Uncertainty? Evaluating Predictive Uncertainty Under Dataset Shift},
  author={Yaniv Ovadia and Emily Fertig and Jie Jessie Ren and Zachary Nado and D. Sculley and Sebastian Nowozin and Joshua V. Dillon and Balaji Lakshminarayanan and Jasper Snoek},
  booktitle={Neural Information Processing Systems},
  year={2019},
  url={https://api.semanticscholar.org/CorpusID:174803437}
}

@inproceedings{blundell2015weight,
author = {Blundell, Charles and Cornebise, Julien and Kavukcuoglu, Koray and Wierstra, Daan},
title = {Weight uncertainty in neural networks},
year = {2015},
publisher = {JMLR.org},
pages = {1613–1622},
numpages = {10},
location = {Lille, France},
booktitle = {Proceedings of the 32nd International Conference on Machine Learning (ICML)},
series = {ICML'15}
}

@inproceedings{louizos2016structured,
author = {Louizos, Christos and Welling, Max},
title = {Structured and efficient variational deep learning with matrix Gaussian posteriors},
year = {2016},
publisher = {JMLR.org},
pages = {1708–1716},
numpages = {9},
location = {New York, NY, USA},
booktitle = {Proceedings of the 33rd International Conference on Machine Learning (ICML)}
}

@inproceedings{hernandez2015probabilistic,
author = {Hern\'{a}ndez-Lobato, Jos\'{e} Miguel and Adams, Ryan P.},
title = {Probabilistic backpropagation for scalable learning of Bayesian neural networks},
year = {2015},
publisher = {JMLR.org},
booktitle = {Proceedings of the 32nd International Conference on International Conference on Machine Learning - Volume 37},
pages = {1861–1869},
numpages = {9},
location = {Lille, France},
series = {ICML'15}
}

@inproceedings{gal2016dropout,
author = {Gal, Yarin and Ghahramani, Zoubin},
title = {Dropout as a Bayesian approximation: representing model uncertainty in deep learning},
year = {2016},
publisher = {JMLR.org},
booktitle = {Proceedings of the 33rd International Conference on International Conference on Machine Learning - Volume 48},
pages = {1050–1059},
numpages = {10},
location = {New York, NY, USA},
series = {ICML'16}
}

@inproceedings{wenzel2020hyperparameter,
author = {Wenzel, Florian and Snoek, Jasper and Tran, Dustin and Jenatton, Rodolphe},
title = {Hyperparameter ensembles for robustness and uncertainty quantification},
year = {2020},
isbn = {9781713829546},
publisher = {Curran Associates Inc.},
address = {Red Hook, NY, USA},
booktitle = {Proceedings of the 34th International Conference on Neural Information Processing Systems},
articleno = {546},
numpages = {14},
location = {Vancouver, BC, Canada},
series = {NIPS '20}
}

@inproceedings{lakshminarayanan2016simple,
author = {Lakshminarayanan, Balaji and Pritzel, Alexander and Blundell, Charles},
title = {Simple and scalable predictive uncertainty estimation using deep ensembles},
year = {2017},
isbn = {9781510860964},
publisher = {Curran Associates Inc.},
address = {Red Hook, NY, USA},
booktitle = {Proceedings of the 31st International Conference on Neural Information Processing Systems},
pages = {6405–6416},
numpages = {12},
location = {Long Beach, California, USA},
series = {NIPS'17}
}

@inproceedings{larrazabal2021orthogonal,
author = {Larrazabal, Agostina J. and Mart\'{\i}nez, C\'{e}sar and Dolz, Jose and Ferrante, Enzo},
title = {Orthogonal Ensemble Networks for Biomedical Image Segmentation},
year = {2021},
isbn = {978-3-030-87198-7},
publisher = {Springer-Verlag},
address = {Berlin, Heidelberg},
url = {https://doi.org/10.1007/978-3-030-87199-4_56},
doi = {10.1007/978-3-030-87199-4_56},
booktitle = {Medical Image Computing and Computer Assisted Intervention – MICCAI 2021: 24th International Conference, Strasbourg, France, September 27–October 1, 2021, Proceedings, Part III},
pages = {594–603},
numpages = {10},
location = {Strasbourg, France}
}

@InProceedings{eace,
author = {Nixon, Jeremy and Dusenberry, Michael W. and Zhang, Linchuan and Jerfel, Ghassen and Tran, Dustin},
title = {Measuring Calibration in Deep Learning},
booktitle = {Proceedings of the IEEE/CVF Conference on Computer Vision and Pattern Recognition (CVPR) Workshops},
month = {June},
year = {2019}
}

@inproceedings{ece,
author = {Naeini, Mahdi Pakdaman and Cooper, Gregory F. and Hauskrecht, Milos},
title = {Obtaining well calibrated probabilities using bayesian binning},
year = {2015},
isbn = {0262511290},
publisher = {AAAI Press},
booktitle = {Proceedings of the Twenty-Ninth AAAI Conference on Artificial Intelligence},
pages = {2901–2907},
numpages = {7},
location = {Austin, Texas},
series = {AAAI'15}
}

@INPROCEEDINGS {mbls,
author = { Liu, Bingyuan and Ayed, Ismail Ben and Galdran, Adrian and Dolz, Jose },
booktitle = { 2022 IEEE/CVF Conference on Computer Vision and Pattern Recognition (CVPR) },
title = {{ The Devil is in the Margin: Margin-based Label Smoothing for Network Calibration }},
year = {2022},
volume = {},
ISSN = {},
pages = {80-88},
doi = {10.1109/CVPR52688.2022.00018},
url = {https://doi.ieeecomputersociety.org/10.1109/CVPR52688.2022.00018},
publisher = {IEEE Computer Society},
address = {Los Alamitos, CA, USA},
month =Jun}

@INPROCEEDINGS{resnet,
  author={He, Kaiming and Zhang, Xiangyu and Ren, Shaoqing and Sun, Jian},
  booktitle={2016 IEEE Conference on Computer Vision and Pattern Recognition (CVPR)}, 
  title={Deep Residual Learning for Image Recognition}, 
  year={2016},
  volume={},
  number={},
  pages={770-778},
  doi={10.1109/CVPR.2016.90}}

@inproceedings{mmca,
  author = {Kumar, Aviral and Sarawagi, Sunita and Jain, Ujjwal},
  booktitle = {ICML},
  editor = {Dy, Jennifer G. and Krause, Andreas},
  ee = {http://proceedings.mlr.press/v80/kumar18a.html},
  pages = {2810-2819},
  publisher = {PMLR},
  series = {Proceedings of Machine Learning Research},
  title = {Trainable Calibration Measures For Neural Networks From Kernel Mean Embeddings.},
  url = {http://dblp.uni-trier.de/db/conf/icml/icml2018.html#KumarSJ18},
  volume = 80,
  year = 2018
}

@article{ecp,
  title={Regularizing Neural Networks by Penalizing Confident Output Distributions},
  author={Gabriel Pereyra and G. Tucker and Jan Chorowski and Lukasz Kaiser and Geoffrey E. Hinton},
  journal={ArXiv},
  year={2017},
  volume={abs/1701.06548},
  url={https://api.semanticscholar.org/CorpusID:9545399}
}

@inproceedings{ls,
author = {M\"{u}ller, Rafael and Kornblith, Simon and Hinton, Geoffrey},
title = {When does label smoothing help?},
year = {2019},
publisher = {Curran Associates Inc.},
address = {Red Hook, NY, USA},
booktitle = {Proceedings of the 33rd International Conference on Neural Information Processing Systems},
articleno = {422},
numpages = {10}
}

@ARTICLE{focalloss,
  author={Lin, Tsung-Yi and Goyal, Priya and Girshick, Ross and He, Kaiming and Dollár, Piotr},
  journal={IEEE Transactions on Pattern Analysis and Machine Intelligence}, 
  title={Focal Loss for Dense Object Detection}, 
  year={2020},
  volume={42},
  number={2},
  pages={318-327},
  doi={10.1109/TPAMI.2018.2858826}}

@inproceedings{ce,
author = {Mao, Anqi and Mohri, Mehryar and Zhong, Yutao},
title = {Cross-entropy loss functions: theoretical analysis and applications},
year = {2023},
publisher = {JMLR.org},
booktitle = {Proceedings of the 40th International Conference on Machine Learning},
articleno = {992},
numpages = {26},
location = {Honolulu, Hawaii, USA},
series = {ICML'23}
}

@INPROCEEDINGS{adaface,
  author={Kim, Minchul and Jain, Anil K. and Liu, Xiaoming},
  booktitle={2022 IEEE/CVF Conference on Computer Vision and Pattern Recognition (CVPR)}, 
  title={AdaFace: Quality Adaptive Margin for Face Recognition}, 
  year={2022},
  volume={},
  number={},
  pages={18729-18738},
  doi={10.1109/CVPR52688.2022.01819}}

@INPROCEEDINGS {vonmisesloss,
author = { Scott, Tyler R. and Gallagher, Andrew C. and Mozer, Michael C. },
booktitle = { 2021 IEEE/CVF International Conference on Computer Vision (ICCV) },
title = {{ von Mises–Fisher Loss: An Exploration of Embedding Geometries for Supervised Learning }},
year = {2021},
volume = {},
ISSN = {},
pages = {10592-10602},
doi = {10.1109/ICCV48922.2021.01044},
url = {https://doi.ieeecomputersociety.org/10.1109/ICCV48922.2021.01044},
publisher = {IEEE Computer Society},
address = {Los Alamitos, CA, USA},
month =Oct}

@inproceedings{adafocal,
author = {Ghosh, Arindam and Schaaf, Thomas and Gormley, Matt},
title = {AdaFocal: calibration-aware adaptive focal loss},
year = {2022},
isbn = {9781713871088},
publisher = {Curran Associates Inc.},
address = {Red Hook, NY, USA},
booktitle = {Proceedings of the 36th International Conference on Neural Information Processing Systems},
articleno = {116},
numpages = {13},
location = {New Orleans, LA, USA},
series = {NIPS '22}
}

@inproceedings{Tao2025FeatureClipping,
  title        = {Feature Clipping for Uncertainty Calibration},
  author       = {Linwei Tao and Minjing Dong and Chang Xu},
  booktitle    = {Proceedings of the AAAI Conference on Artificial Intelligence},
  year         = {2025},
  volume       = {39},
  pages        = {20841--20849},
  publisher    = {AAAI Press},
  doi          = {10.1609/aaai.v39i19.34297},
  url          = {https://ojs.aaai.org/index.php/AAAI/article/view/34297}
}

@inproceedings{prob,
author = {Yang, Jia-Qi and Zhan, De-Chuan and Gan, Le},
title = {Beyond probability partitions: calibrating neural networks with semantic aware grouping},
year = {2023},
publisher = {Curran Associates Inc.},
address = {Red Hook, NY, USA},
booktitle = {Proceedings of the 37th International Conference on Neural Information Processing Systems},
articleno = {2547},
numpages = {13},
location = {New Orleans, LA, USA},
series = {NIPS '23}
}

@inproceedings{HbertJohnson2018MulticalibrationCF,
  title={Multicalibration: Calibration for the (Computationally-Identifiable) Masses},
  author={{\'U}rsula H{\'e}bert-Johnson and Michael P. Kim and Omer Reingold and Guy N. Rothblum},
  booktitle={International Conference on Machine Learning},
  year={2018},
  url={https://api.semanticscholar.org/CorpusID:51880858}
}

@INPROCEEDINGS{rankmixup,
  author={Noh, Jongyoun and Park, Hyekang and Lee, Junghyup and Ham, Bumsub},
  booktitle={2023 IEEE/CVF International Conference on Computer Vision (ICCV)}, 
  title={RankMixup: Ranking-Based Mixup Training for Network Calibration}, 
  year={2023},
  volume={},
  number={},
  pages={1358-1368},
  doi={10.1109/ICCV51070.2023.00131}}

@article{ste,
  title={Estimating or Propagating Gradients Through Stochastic Neurons for Conditional Computation},
  author={Yoshua Bengio and Nicholas L{\'e}onard and Aaron C. Courville},
  journal={ArXiv},
  year={2013},
  volume={abs/1308.3432},
  url={https://api.semanticscholar.org/CorpusID:18406556}
}

@ARTICLE{ood,
  author={De Bernardi, Giacomo and Narteni, Sara and Cambiaso, Enrico and Mongelli, Maurizio},
  journal={IEEE Transactions on Artificial Intelligence}, 
  title={Rule-Based Out-of-Distribution Detection}, 
  year={2024},
  volume={5},
  number={6},
  pages={2627-2637},
  doi={10.1109/TAI.2023.3323923}}

@inproceedings{CRLcheck,
author = {Moon, Jooyoung and Kim, Jihyo and Shin, Younghak and Hwang, Sangheum},
title = {Confidence-aware learning for deep neural networks},
year = {2020},
publisher = {JMLR.org},
booktitle = {Proceedings of the 37th International Conference on Machine Learning},
articleno = {652},
numpages = {11},
series = {ICML'20}
}

@INPROCEEDINGS{acls,
  author={Park, Hyekang and Noh, Jongyoun and Oh, Youngmin and Baek, Donghyeon and Ham, Bumsub},
  booktitle={2023 IEEE/CVF International Conference on Computer Vision (ICCV)}, 
  title={ACLS: Adaptive and Conditional Label Smoothing for Network Calibration}, 
  year={2023},
  volume={},
  number={},
  pages={3913-3922},
  doi={10.1109/ICCV51070.2023.00364}}

@inproceedings{dfl,
author = {Tao, Linwei and Dong, Minjing and Xu, Chang},
title = {Dual focal loss for calibration},
year = {2023},
publisher = {JMLR.org},
booktitle = {Proceedings of the 40th International Conference on Machine Learning},
articleno = {1410},
numpages = {17},
location = {Honolulu, Hawaii, USA},
series = {ICML'23}
}

@InProceedings{vaze,
               title={Open-Set Recognition: a Good Closed-Set Classifier is All You Need?},
               author={Sagar Vaze and Kai Han and Andrea Vedaldi and Andrew Zisserman},
               booktitle={International Conference on Learning Representations},
               year={2022}}

@InProceedings{sphor,
    author    = {Bahavan, Thiru Thillai Nadarasar and Seneviratne, Sachith and Halgamuge, Saman},
    title     = {SpHOR: A Representation Learning Perspective on Open-set Recognition for Identifying Unknown Classes in Deep Neural Networks},
    booktitle = {Proceedings of the IEEE/CVF Conference on Computer Vision and Pattern Recognition (CVPR) Findings},
    month     = {June},
    year      = {2026},
    pages     = {6901-6910}
}

@article{Maji2013FineGrained,
  author    = {Maji, Subhransu and Rahtu, Esa and Kannala, Juho and Blaschko, Matthew and Vedaldi, Andrea},
  title     = {Fine-Grained Visual Classification of Aircraft},
  journal   = {arXiv preprint arXiv:1306.5151},
  year      = {2013},
  url       = {https://arxiv.org/abs/1306.5151}
}

@inproceedings{ViT_ICLR2021dosovitskiy,
  author = {Dosovitskiy, Alexey and Beyer, Lucas and Kolesnikov, Alexander and Weissenborn, Dirk and Zhai, Xiaohua and Unterthiner, Thomas and Dehghani, Mostafa and Minderer, Matthias and Heigold, Georg and Gelly, Sylvain and Uszkoreit, Jakob and Houlsby, Neil},
  booktitle = {International Conference on Learning Representations},
  title = {An Image is Worth 16x16 Words: Transformers for Image Recognition at Scale},
  year = 2021
}

@inproceedings{decoupling,
author = {Jordahn, Mikkel and Olmos, Pablo M.},
title = {Decoupling feature extraction and classification layers for calibrated neural networks},
year = {2024},
publisher = {JMLR.org},
booktitle = {Proceedings of the 41st International Conference on Machine Learning},
articleno = {905},
numpages = {21},
location = {Vienna, Austria},
series = {ICML'24}
}

@inproceedings{calibfocal,
author = {Mukhoti, Jishnu and Kulharia, Viveka and Sanyal, Amartya and Golodetz, Stuart and Torr, Philip H. S. and Dokania, Puneet K.},
title = {Calibrating deep neural networks using focal loss},
year = {2020},
isbn = {9781713829546},
publisher = {Curran Associates Inc.},
address = {Red Hook, NY, USA},
booktitle = {Proceedings of the 34th International Conference on Neural Information Processing Systems},
articleno = {1282},
numpages = {12},
location = {Vancouver, BC, Canada},
series = {NIPS '20}
}

@INPROCEEDINGS{CPC,
  author={Cheng, Jiacheng and Vasconcelos, Nuno},
  booktitle={2022 IEEE/CVF Conference on Computer Vision and Pattern Recognition (CVPR)}, 
  title={Calibrating Deep Neural Networks by Pairwise Constraints}, 
  year={2022},
  volume={},
  number={},
  pages={13699-13708},
  doi={10.1109/CVPR52688.2022.01334}}

@InProceedings{MDCA,
    author    = {Hebbalaguppe, Ramya and Prakash, Jatin and Madan, Neelabh and Arora, Chetan},
    title     = {A Stitch in Time Saves Nine: A Train-Time Regularizing Loss for Improved Neural Network Calibration},
    booktitle = {Proceedings of the IEEE/CVF Conference on Computer Vision and Pattern Recognition (CVPR)},
    month     = {June},
    year      = {2022},
    pages     = {16081-16090}
}

\end{document}